\documentclass[a4paper,fleqn]{cas-sc}

\usepackage[english]{babel}
\usepackage{array}
\usepackage{multirow}
\usepackage{amsmath}
\usepackage{graphicx}
\usepackage{booktabs}
\usepackage{adjustbox}
\usepackage{longtable}
\usepackage{placeins}
\usepackage[authoryear]{natbib}

\begin{document}

\let\WriteBookmarks\relax
\def\floatpagepagefraction{1}
\def\textpagefraction{.001}

\shorttitle{Transformer-Based Autonomous Driving Models and Deployment-Oriented Compression}
\shortauthors{J. Zhong et al.}

\title[mode=title]{Transformer-Based Autonomous Driving Models and Deployment-Oriented Compression: A Survey}

\author[1]{Juan Zhong}
\fnmark[1]

\author[2,3]{Yuhang Shi}
\fnmark[1]

\author[4]{Zukang Xu}
\ead{zukang.xu@houmo.ai}

\author[2,3]{Xi Chen}
\cormark[1]
\ead{x_chen@fudan.edu.cn}

\affiliation[1]{
    organization={Renmin University of China},
    city={Beijing},
    country={China}
}

\affiliation[2]{
    organization={Artificial Intelligence Innovation and Incubation Institute, Fudan University},
    city={Shanghai},
    country={China}
}

\affiliation[3]{
    organization={Shanghai Academy of AI for Science},
    city={Shanghai},
    country={China}
}

\affiliation[4]{
    organization={Department of houmo.ai}
}

\cortext[cor1]{Corresponding author}
\fntext[fn1]{Yuhang Shi and Juan Zhong contributed equally to this work.}

\begin{abstract}
Transformer-based models are becoming a central paradigm in autonomous driving because they can capture long-range spatial dependencies, multi-agent interactions, and multimodal context across perception, prediction, and planning. At the same time, their deployment in real vehicles remains difficult because high-capacity attention-based architectures impose substantial latency, memory, and energy overhead. This survey reviews representative Transformer-based autonomous driving models and organizes them by task role, sensing configuration, and architectural design. More importantly, it examines these models from a deployment-oriented perspective and analyzes how efficiency constraints reshape model design choices in practice. We further review compression and acceleration strategies relevant to Transformer-based driving systems, including quantization, pruning, knowledge distillation, low-rank approximation, and efficient attention, and discuss their benefits, limitations, and task-dependent applicability. Rather than treating compression as an isolated post-processing step, we highlight it as a system-level design consideration that directly affects deployability, robustness, and safety. Finally, we identify open challenges and future research directions toward standardized, safety-aware, and hardware-conscious evaluation of efficient autonomous driving systems.
\end{abstract}


\maketitle

\section{Introduction}

\begin{figure}
\centering
\includegraphics[width=0.95\textwidth]{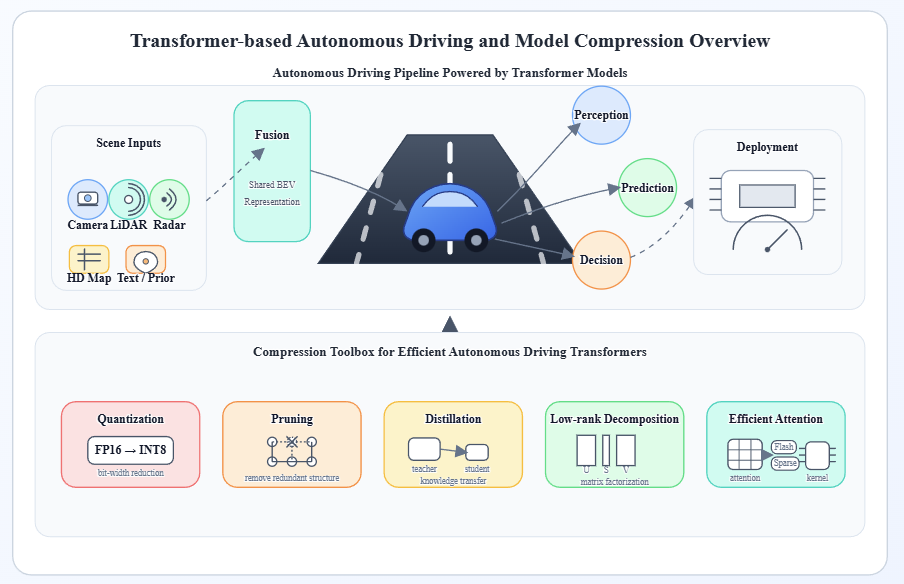}
\caption{\label{fig:overview} Overview of this survey. Transformer-based autonomous driving systems are organized from multimodal inputs and shared representations to downstream tasks, while deployment constraints motivate compression strategies such as quantization, pruning, distillation, low-rank decomposition, and efficient attention.}
\end{figure}

Autonomous driving has emerged as a transformative technology with the potential to improve transportation safety, efficiency, and accessibility through increasingly capable automated systems. The field's technical evolution spans two decades, progressing from rule-based algorithms~\citep{bengler2014three, hussain2018autonomous} to data-driven machine learning approaches~\citep{grigorescu2020survey}. Traditional approaches for autonomous driving mainly employed analytical formulations and encoded traffic rules for environmental perception and vehicle control, as evidenced by foundational works in urban scenarios~\citep{campbell2010autonomous}, motion planning and control~\citep{katrakazas2015real, paden2016survey}, and simultaneous localization and mapping (SLAM) tasks~\citep{bresson2017simultaneous}. However, these traditional methods often face challenges in handling complex real-world scenarios, such as dynamic objects, occlusions, and uncertain environments. In contrast, deep learning methods~\citep{lecun2015deep}, particularly deep neural networks (DNNs), have demonstrated remarkable performance in learning complex patterns from data and making predictions. For instance, convolutional neural networks (CNNs)~\citep{gu2018recent}, a representative class of DNNs, utilize convolutional layers to extract local spatial features and hierarchically combine them to model complex visual patterns. As a result, CNN-based models have been widely adopted in various autonomous driving applications. Recent YOLO-based studies further indicate that CNN-based detectors remain competitive for efficient object-level perception in autonomous driving by incorporating lightweight backbones, attention mechanism, and improved bounding-box regression losses~\citep{guo2024research}.Deep learning methods for different autonomous driving tasks have been reviewed and discussed in previous surveys, including general discussions~\citep{grigorescu2020survey, yurtsever2020survey, kuutti2020survey, bachute2021autonomous, parekh2022review}; models using reinforcement learning~\citep{kiran2021deep}; models for object detection~\citep{alaba2023deep}, trajectory and behavior prediction~\citep{mozaffari2020deep, huang2022survey}, multi-modal fusion~\citep{cui2021deep}, planning and decision making~\citep{schwarting2018planning, liu2021decision, abdallaoui2022thorough}, explainability AI~\citep{zablocki2022explainability}, and scenario generation~\citep{ding2023survey}. 

Nevertheless, modern autonomous driving systems must simultaneously process heterogeneous sensor inputs, reason over long-range spatial dependencies, and model temporal interactions among multiple agents. These requirements expose the limitations of convolution-dominated architectures, particularly in scenarios involving complex multi-modal fusion and long-term spatiotemporal reasoning.

In parallel with the advancement of deep learning, Transformer architectures~\citep{vaswani2017attention} have demonstrated strong performance across a wide range of autonomous driving tasks, including perception, prediction, and planning, largely due to their capacity to model long-range dependencies and complex interactions. Consequently, a growing body of research has explored Transformer-based model designs tailored to different driving subtasks, sensor modalities, and system configurations. This survey reviews these Transformer-based autonomous driving models through a deployment-oriented lens and analyzes the practical challenges that arise under real-time and resource-constrained operating conditions.

The architectural advantages of Transformers originate from attention mechanisms~\citep{bahdanau2015neural}, which enable models to selectively focus on relevant information based on contextual relevance and capture global dependencies more effectively than conventional convolutional or recurrent architectures. The Transformer architecture~\citep{vaswani2017attention}, originally proposed for natural language processing, has been increasingly adopted in autonomous driving due to its flexibility in modeling multi-modal and sequential data. These properties have motivated the widespread adoption of Transformers for modeling heterogeneous and sequential driving data in autonomous driving systems.

Despite their representational advantages, Transformer-based models typically involve large parameter sizes and high computational complexity, posing significant challenges for real-time deployment on onboard platforms with strict latency, memory, and power constraints. While architectural optimizations—such as scaling down model size—offer partial relief, they often necessitate a compromise between inference speed and representational capacity, thereby limiting robustness in complex driving scenarios. Accordingly, model compression has evolved from a secondary optimization into a central design consideration for reconciling high-capacity modeling with resource-constrained deployment. By leveraging techniques such as quantization, pruning, knowledge distillation, and low-rank decomposition, it is possible to drastically reduce computational overhead while preserving the task-critical accuracy essential for safe autonomous driving systems.

Although several existing surveys have reviewed Transformer architectures or deep learning techniques for autonomous driving, most primarily emphasize model families, task taxonomies, or application scenarios, with comparatively limited attention to deployment-oriented efficiency constraints. A recent broad survey by \citet{chu2025survey}, for example, provides a comprehensive overview of Transformer architectures in autonomous driving from the perspectives of task taxonomy, industrial practice, and emerging large-model directions. That line of work is valuable for understanding architectural evolution, but it leaves a practical gap that is increasingly important for real-world deployment: how Transformer design choices interact with latency, memory, power, and compression requirements across different tasks in the driving stack. In particular, the relationship between task characteristics and compression behavior remains insufficiently synthesized. Compression decisions that are acceptable for map construction or lane detection may be unsafe or ineffective for multi-view 3D detection, multimodal fusion, or closed-loop planning. A focused survey at this intersection is therefore needed, not merely to catalogue models, but to clarify which efficiency strategies are plausible, risky, or still underexplored in safety-critical autonomous driving systems.

To address this gap, this survey provides a systematic review of 
Transformer-based autonomous driving models together with an in-depth 
analysis of model compression techniques applied in this domain. Rather than offering another broad overview of Transformer applications, this paper specifically focuses on the intersection between task-level model design and deployment-oriented compression, aiming to clarify how efficiency constraints reshape model selection, optimization strategy, and real-world applicability across autonomous driving tasks. 
Against this background, the main contributions of this survey are as follows.
\begin{enumerate}
    \item It synthesizes representative Transformer-based architectures across the major autonomous driving tasks, including perception, prediction, planning, and end-to-end systems.
    \item It reviews model compression methods relevant to Transformer-based autonomous driving models and analyzes their task-dependent characteristics, benefits, and limitations.
    \item It discusses the trade-offs between efficiency, accuracy, and deployability under realistic system constraints, and highlights future research directions toward practical deployment.
\end{enumerate}
Figure~\ref{fig:overview} provides a high-level overview of the survey scope. It connects multimodal driving inputs, shared representations, downstream driving tasks, deployment constraints, and the main compression strategies discussed in the following sections.

With this overall framing in place, the remainder of this paper is organized as follows. Section~\ref{Sec: models} reviews Transformer-based model architectures and system designs for different autonomous driving tasks. Section~\ref{Sec: compression} summarizes model compression techniques that have been applied to Transformer-based autonomous driving systems and analyzes their characteristics and limitations. Section~\ref{Sec: Conclusion} discusses open challenges and future research directions, followed by conclusions.  

\begin{figure}
\centering
\includegraphics[width=0.8\textwidth]{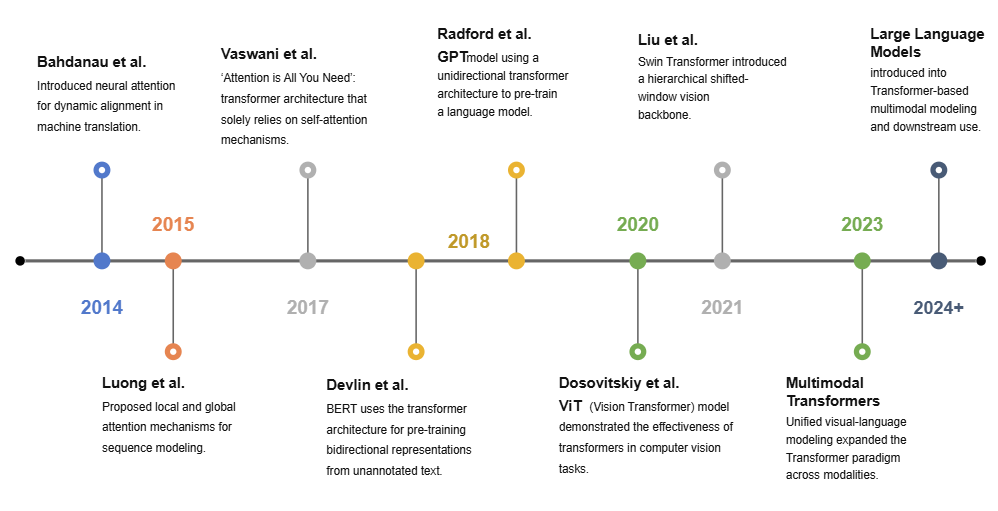}
\caption{\label{fig:history} A timeline diagram illustrating the history and key milestones of attention mechanisms and Transformer architectures research.}
\end{figure}

\begin{figure}
\centering
\includegraphics[width=0.5\textwidth]{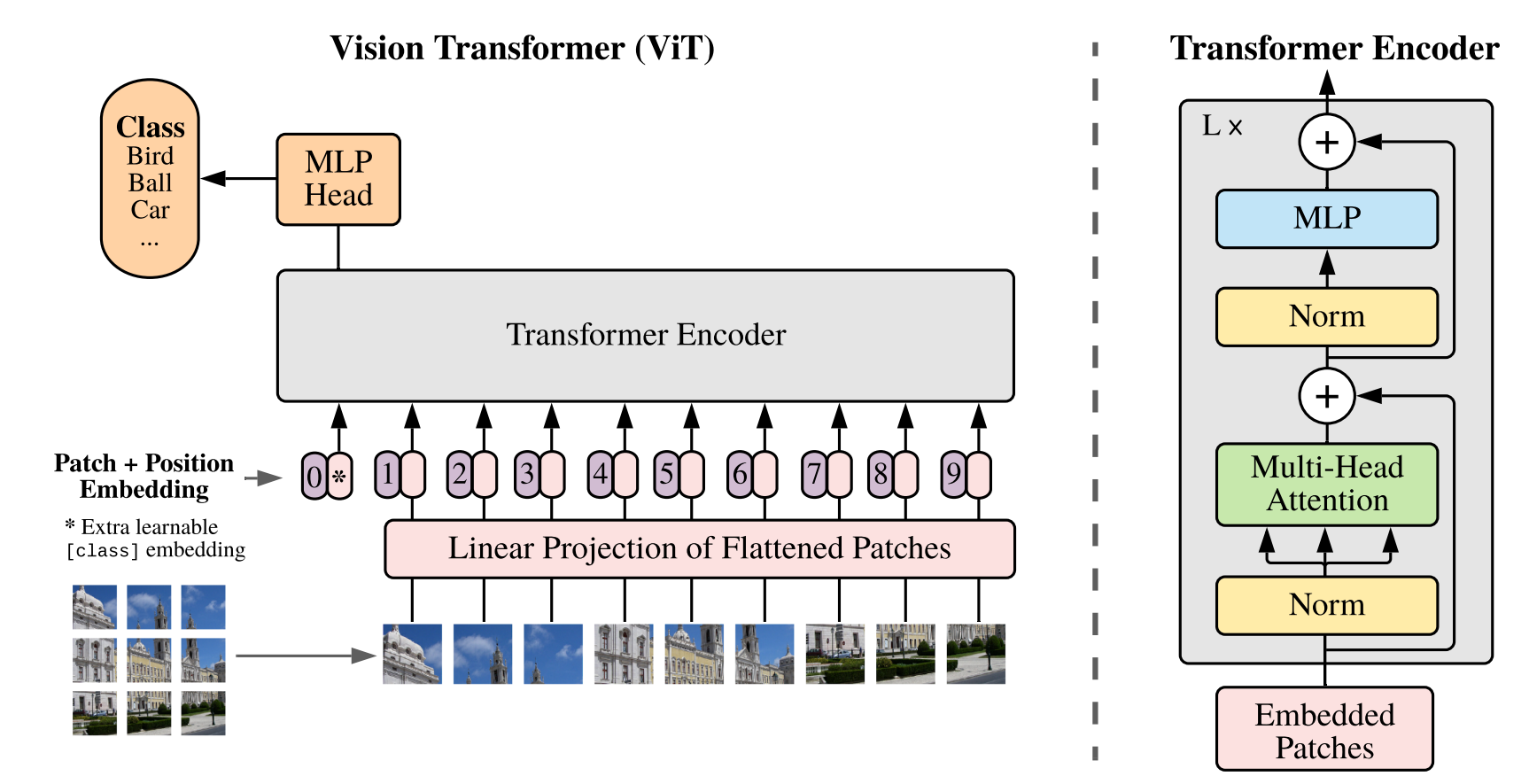}
\caption{\label{fig:vit} The architecture of ViT, the left panel shows the image division and position embedding process and the right panel presents a standard encoder architecture that contains the multi-head attention layer, as illustrated in~\citep{dosovitskiy2020image}. }
\end{figure}

\section{Transformer Models and Tasks}
\label{Sec: models}

Figure~\ref{fig:history} summarizes major milestones in the development of attention mechanisms and Transformer architectures that underpin modern vision and autonomous driving research. \citet{bahdanau2015neural} first introduced the attention mechanism in the context of neural machine translation, proposing a dynamic alignment method between source and target sequences. This approach overcame the limitations of fixed-length context vectors in earlier sequence-to-sequence models. \citet{luong2015effective} further refined attention mechanisms by presenting local and global attention, with the former focusing on a smaller source sequence subset and the latter considering all source words for variable-length alignment context computation.

A milestone in this line of research is the work of \citet{vaswani2017attention}, which introduced the Transformer architecture and demonstrated the effectiveness of attention-centric sequence modeling. This innovation led to substantial performance gains across a wide range of NLP tasks. Building upon this, \citet{devlin2019bert} proposed BERT (Bidirectional Encoder Representations from Transformers), a pre-trained model for bidirectional representations using the Transformer architecture. When fine-tuned for downstream tasks, BERT achieved unprecedented performance in NLP. Concurrently, \citet{radford2018improving} presented the GPT (Generative Pre-trained Transformer) model, which employed a unidirectional Transformer architecture for language model pre-training. Fine-tuning GPT on specific tasks yielded substantial performance improvements, and later iterations (GPT-2, GPT-3, and GPT-4) continued to advance the state of the art. \citet{dosovitskiy2020image} demonstrated the applicability of Transformer architecture to computer vision tasks with the Vision Transformer (ViT) model. By dividing images into non-overlapping patches and using linear embeddings, the authors achieved competitive results compared to traditional CNN models in image classification tasks. ViT also became a foundational backbone for many subsequent Transformer-based models in image understanding and autonomous driving. 

\begin{table}[!ht]
\centering
\caption{\label{tab:modelTable} Representative Transformer-based models organized by autonomous driving task, including object detection and tracking, 3D segmentation, lane detection and segmentation, high-definition (HD) map generation, trajectory and behavior prediction, and end-to-end driving. The primary sensor modalities are RGB cameras and LiDAR, while HD map information is additionally used in trajectory prediction settings. `BEV' indicates whether the model produces bird's-eye-view features. } 

\footnotesize
\setlength{\tabcolsep}{4pt}
\begin{adjustbox}{max width=\textwidth}
\begin{tabular}{@{}llllc@{}}

\toprule
\textbf{Model} & \textbf{Tasks} & \textbf{Data Sources} & \textbf{BEV} & \textbf{Release Year} \\
\midrule
\multicolumn{5}{@{}l}{\textbf{3D \& General Perception}} \\
\midrule
DETR3D~\citep{wang2022detr3d} & Object Detection & Camera & N & 2021 \\
FUTR3D~\citep{chen2023futr3d} & Object Detection & Camera, LiDAR & N & 2022 \\
PETR~\citep{liu2022petr,liu2022petrv2} & Object Detection & Camera & N & 2022 \\
CrossDTR~\citep{tseng2022crossdtr} & Object Detection & Camera & Y & 2022 \\
BEVFormer~\citep{li2022bevformer,chenyu2022bevformerv2} & Object Detection & Camera & Y & 2022 \\
UVTR~\citep{uvtr2022yanwei} & Object Detection & Camera, LiDAR & Y & 2022 \\
M3DETR~\citep{guan2022m3detr} & Object Detection & Camera, LiDAR & Y & 2022 \\
StreamPETR~\citep{wang2023exploring} & Object Detection & Camera & N & 2023 \\
BEVFusion4D~\citep{cai2023bevfusion4d} & Object Detection & Camera, LiDAR & Y & 2024 \\
IS-Fusion~\citep{yin2024fusion} & Object Detection & Camera, LiDAR & Y & 2024 \\
RCTrans~\citep{li2025rctrans} & Object Detection & Camera, LiDAR & Y & 2025 \\
RCBEVDet~\citep{lin2024rcbevdet} & Object Detection & Camera, Radar & Y & 2024 \\
TPVFormer~\citep{tpvformer2023huang} & 3D Segmentation & Camera & Y & 2023 \\
VoxFormer~\citep{yiming2023voxformer} & 3D Segmentation & Camera & Y & 2023 \\
SurroundOCC~\citep{yiwei2023surroundocc} & 3D Segmentation & Camera & Y & 2023 \\
Cam4DOCC~\citep{ma2024cam4docc} & 3D Segmentation & Camera & Y & 2024 \\
MOTR~\citep{motr2022zeng} & Object Tracking & Camera & N & 2022 \\
MUTR3D~\citep{zhang2022mutr3d} & Object Tracking & Camera & N & 2022 \\
MeMOTR~\citep{gao2023memotr} & Object Tracking & Camera & N & 2023 \\
STT~\citep{jing2024stt} & Object Tracking & Camera, LiDAR & N & 2024 \\
TrajectoryFormer~\citep{chen2023trajectoryformer} & Object Tracking & Camera & N & 2023 \\
\midrule
\multicolumn{5}{@{}l}{\textbf{2D \& Plane}} \\
\midrule
BEVSegFormer~\citep{peng2023bevsegformer} & Lane Detection & Camera & Y & 2022 \\
Persformer~\citep{persformer2022lichen} & Lane Detection & Camera & Y & 2022 \\
LSTR~\citep{lstr2021ruijin} & Lane Detection & Camera & N & 2020 \\
CurveFormer~\citep{bai2022curveformer} & Lane Detection & Camera & Y & 2023 \\
CurveFormer++~\citep{bai2025curveformer++} & Lane Detection & Camera & Y & 2025 \\
HWLane~\citep{zhao2024hwlane} & Lane Detection & Camera & N & 2024 \\
LATR~\citep{luo2023latr} & Lane Detection & Camera & N & 2023 \\
TIiM~\citep{tiim2022saha} & Segmentation & Camera & Y & 2022  \\
Panoptic SegFormer~\citep{panoptic2022li} & Segmentation & Camera & N & 2022 \\
% \midrule
STSU~\citep{stsu2021yigit} & HD Map Generation & Camera & N & 2021 \\
VectorMapNet~\citep{liu2023vectormapnet} & HD Map Generation & Camera, LiDAR & N & 2022 \\
MapTR~\citep{maptr2023bencheng} & HD Map Generation & Camera & Y & 2023 \\
StreamMapNet~\citep{yuan2024streammapnet} & HD Map Generation & Camera & Y & 2024 \\
CAFMap~\citep{fang2025cafmap} & HD Map Generation & Camera & Y & 2025 \\
PrevPredMap~\citep{peng2025prevpredmap} & HD Map Generation & Camera & Y & 2025 \\
\midrule
\multicolumn{5}{@{}l}{\textbf{Prediction \& Decision Making}} \\
\midrule
VectorNet~\citep{vectornet2020gao} & Trajectory/Behavior Prediction & Trajectory and Map info  & N & 2020 \\
DenseTNT~\citep{densetnt2021gu} & Trajectory/Behavior Prediction & Trajectory and Map info  & N & 2021 \\
mmTransformer~\citep{mmTransformer2022liu} & Trajectory/Behavior Prediction & Trajectory and Map info  & N & 2021 \\
AgentFormer~\citep{agentformer2021yuan} & Trajectory/Behavior Prediction & Camera & N & 2021  \\
WayFormer~\citep{wayformer2022nayakanti} & Trajectory/Behavior Prediction & Trajectory and Map info  & N & 2022 \\
HPTR~\citep{zhang2023real} & Trajectory/Behavior Prediction & Trajectory and Map info  & N & 2023 \\
Co-MTP~\citep{zhang2025co} & Trajectory/Behavior Prediction & Trajectory and Map info  & N & 2025 \\
IA-TTLP~\citep{liang2025interaction} & Trajectory/Behavior Prediction & Trajectory and Map info & N & 2025 \\
% \midrule
TransFuser~\citep{transfuser2022chitta} & End-to-End & Camera, LiDAR & Y & 2022 \\
NEAT~\citep{neat2021chitta} & End-to-End & Camera & Y & 2021  \\
InterFuser~\citep{interfuser2023shao-safety} & End-to-End & Camera, LiDAR & Y & 2022 \\
MMFN~\citep{zhang2022mmfn} & End-to-End & Camera, LiDAR, Radar, HD Map &  Y & 2022\\
STP3~\citep{stp32022hu} & End-to-End & Camera & Y & 2022 \\
UniAD~\citep{uniad2023hu} & End-to-End & Camera & Y & 2023\\
ReasonNet~\citep{shao2023reasonnet} & End-to-End & Camera, LiDAR &  Y & 2023\\
Scene-Rep Transformer~\citep{liu2024augmenting} & End-to-End & Camera, LiDAR, Radar, HD Map & Y & 2024\\
BridgeAD~\citep{zhang2025bridging} & End-to-End & Camera, LiDAR, Radar, HD Map & Y & 2025\\
DiffusionDrive~\citep{liao2025diffusiondrive} & End-to-End & Camera, LiDAR, Radar, HD Map & Y & 2025\\
DriveGPT4~\citep{xu2024drivegpt4} & Language-augmented Decision Making & Camera, Text & N & 2023\\
VLP~\citep{pan2024vlp} & Language-augmented Decision Making & Camera, Text & N & 2024\\
DriveVLM~\citep{tian2024drivevlm} & Language-augmented Decision Making & Camera, Text & N & 2024\\
DiLU~\citep{wen2023dilu} & Language-augmented Decision Making & simulation environment, Text & N & 2024\\
SimLingo~\citep{renz2025simlingo} & Language-augmented Decision Making & Camera, Text & N & 2025\\
\bottomrule
\end{tabular}
\end{adjustbox}
\end{table}

\FloatBarrier

\subsection{The Base Models for Image Processing}

As a pioneering architecture for Transformer-based image processing, ViT~\citep{dosovitskiy2020image} replaced most convolutional operations in traditional CNN pipelines with self-attention-based token interactions. As shown in Figure~\ref{fig:vit}, an input image is divided into a sequence of non-overlapping patches that are then fed into a Transformer encoder to learn a global representation. The Transformer encoder consists of several self-attention layers followed by feed-forward layers. The self-attention mechanism allows the network to emphasize informative patches while suppressing less relevant ones. To make ViT applicable to larger images, the authors introduced a hybrid approach combining convolutional layers with self-attention layers. The convolutional layers reduce the spatial resolution of the image, while the self-attention layers capture the long-range dependencies between patches.

Inspired by ViT, Swin Transformer~\citep{liu2021swin} introduces a hierarchical architecture that organizes self-attention into multiple stages, each operating on groups of non-overlapping patches. This design addresses the scalability limitations caused by the quadratic complexity of global self-attention. Swin Transformer''s key innovation is shifted-window attention, which enables local cross-window interaction while maintaining manageable computational cost. It also adopts a hierarchical tokenization strategy that recursively groups fine-grained patches into coarser representations, preserving spatial structure across scales and enabling the model to capture both local detail and broader context. Many perception models adopt Swin Transformer as the image backbone, including BEVFusion~\citep{liu2022bevfusion,liang2022bevfusion} and BEVerse~\citep{zhang2022beverse}.

In autonomous driving, Transformer-based architectures have been adopted across a wide range of sub-tasks, including object detection, lane detection and segmentation, tracking and localization, path planning, and decision-making. Recent studies explore the use of Transformer for constructing end-to-end deep learning models for autonomous driving. These models leverage the attention mechanism to further improve their ability to focus on relevant information and perform effectively in complex, real-world driving scenarios. In the remainder of this section, we review representative Transformer-based models according to their task roles, as summarized in Table~\ref{tab:modelTable}. We primarily divide the tasks into three categories: 3D \& general perception tasks (including object detection, tracking, and 3D segmentation); 2D \& plane tasks (including lane detection, segmentation, and High Definition (HD) map generation); and other tasks (including trajectory prediction, behavior prediction, and end-to-end tasks). 

\subsection{3D \& General Perception Tasks}
The first category comprises 3D and general perception tasks, including object detection, tracking, and 3D segmentation. This domain has emerged as a focal point for Transformer-based model development in recent years, aiming to segment, identify, and track objects such as vehicles, pedestrians, and other elements in the environment. Amongst various Transformer-based models, DETR~\citep{carion2020end} is an early important one that inspired many subsequent works, though originally designed for 2D detection. DETR views object detection as a direct set prediction problem. It uses a fixed set of learned object queries and removes hand-designed components such as anchor generation and NMS via bipartite matching. To address DETR's limitations in convergence speed and query ambiguity, Deformable DETR\cite{xizhou2020deformabledetr} introduces adaptive attention mechanisms through deformable queries. For 3D perception, DETR3D~\citep{wang2022detr3d} extends this set-prediction paradigm to multi-camera setups in a camera-only manner. Instead of relying on LiDAR point clouds or explicit dense depth reconstruction, it extracts 2D features from multi-view images and uses sparse 3D queries to index image features through camera projection matrices. FUTR~\citep{gong2022future} follows a similar architecture to DETR but utilizes multi-sensor inputs (images+LiDAR+radar). Multi-modal inputs are fused into a BEV feature and lifted to produce 3D bounding boxes. Building on FUTR, FUTR3D~\citep{chen2023futr3d} extends 3D object detection to multimodal fusion. Similar to DETR3D in structure, it adds a Modality-Agnostic Feature Sampler (MAFS), capable of handling various sensor configurations and fusing different modalities, including 2D cameras, 3D LiDARs, 3D radars, and 4D imaging radars. IS-Fusion~\citep{yin2024fusion} comprises an innovative multimodal fusion framework capturing both instance-level and scene-level contextual information. The method comprises two key modules: the Hierarchical Scene Fusion (HSF) module and the Instance-Guided Fusion (IGF) module. The HSF module uses Point-to-Grid and Grid-to-Region transformers to capture scene features. The IGF module establishes instance-scene relationships through attention mechanisms. By integrating local instance features with global semantic context, IS-Fusion generates BEV representations suitable for instance-aware tasks. To address the challenges of sparsity and noise in radar point clouds for radar-camera 3D object detection, RCTrans~\citep{li2025rctrans} introduces a novel query-based detection method. It enriches sparse valid radar tokens using a Radar Dense Encoder and progressively fuses multimodal features to predict 3D bounding boxes through a Pruning Sequential Decoder.

Despite these advances, existing camera-centric BEV pipelines still suffer from reduced reliability in adverse weather and distant-range perception. Recent radar-camera BEV detection methods therefore further strengthen this research line. RCBEVDet~\citep{lin2024rcbevdet} explicitly exploits radar geometric cues and camera semantic cues in a unified BEV representation, improving robustness in low-visibility and long-range scenarios where image-only perception is fragile. Taken together with RCTrans, this trend suggests that Transformer-based cross-modal fusion is moving from generic feature concatenation to modality-aware interaction design.

PETR~\citep{liu2022petr,liu2022petrv2} is another recent development using a position embedding transformation for multi-view 3D object detection. It encodes 3D coordinate position information into image features, producing 3D position-aware features. During inference, 3D position coordinates can be generated offline and used as additional position embeddings. Building on this foundation, PETRv2 extends the 3D position embedding in PETR for temporal modeling. It utilizes the temporal information of previous frames to boost 3D object detection and support multitask learning (e.g., BEV segmentation and 3D lane detection) by introducing task-specific queries. Meanwhile, StreamPETR~\citep{wang2023exploring} further advances temporal modeling in PETRv2 by introducing an object-centric temporal mechanism. It efficiently propagates historical information through object queries frame by frame, allowing the model to capture long-term spatial-temporal interactions while maintaining high efficiency. CrossDTR~\citep{tseng2022crossdtr}, which combines the strengths of PETR and DETR3D, creates a cross-view and depth-guided framework that achieves accuracy comparable to other methods while enabling faster processing through fewer decoder layers. 
BEVFormer~\citep{li2022bevformer,chenyu2022bevformerv2} employs a different approach using spatio-temporal Transformer architecture for unified BEV representations to improve performance without relying on multi-modal inputs. It incorporates both spatial and temporal fusion, leveraging historical information for enhanced performance. BEVFormer employs a temporal self-attention module to extract features from historical BEV features for motion target velocity estimation and occluded target detection, and spatial cross-attention is extended in the vertical direction for columnar queries in the Z direction of BEV. In contrast, UVTR~\citep{uvtr2022yanwei} focuses on enhancing depth inference by using cross-modal interaction between image and LiDAR inputs, generating separate voxel spaces for each modality in BEV without height compression, and then fusing the multimodal information via knowledge transfer and modal fusion. This approach provides a promising direction for expanding 3D occupancy research. M3DETR~\citep{guan2022m3detr} is a novel Transformer-based architecture for 3D object detection, integrating multiple point cloud representations (raw point clouds, voxels, bird’s-eye-view) and multi-scale features while modeling the mutual relationships between point clouds. The model effectively fuses features across different scales and representations using multi-representation, multi-scale, and mutual-relation Transformers, and it performs 3D object detection through a Region Proposal Network (RPN) and a Region-based Convolutional Neural Network (R-CNN).

Recently, BEVFusion4D~\citep{cai2023bevfusion4d} emerges as a state-of-the-art method integrating LiDAR and camera information into BEV space through cross-modality guidance and temporal aggregation. Unlike previous methods using independent dual-branch frameworks for separate LiDAR/camera BEV feature generation, BEVFusion4D proposes a LiDAR-Guided View Transformer (LGVT) to obtain camera representations in BEV space. The LGVT treats camera BEV as primitive semantic queries, repeatedly leveraging LiDAR BEV's spatial cues to extract image features across camera views. Furthermore, BEVFusion4D extends into the temporal domain with a Temporal Deformable Alignment (TDA) module aggregating historical BEV features to capture motion cues and enhance detection accuracy.

In 3D segmentation, TPVFormer~\citep{tpvformer2023huang} addresses efficiency issues by transforming volumes into three BEV planes, significantly reducing computational burden while effectively predicting semantic occupancy for all voxels. VoxFormer~\citep{yiming2023voxformer} uses 2D images to generate 3D voxel query proposals through depth prediction, then performs deformable cross-attention queries from 2D image features based on these proposals. It subsequently applies a masked autoencoder to propagate information via self-attention and refines voxels through an upsample network to generate semantic occupancy results. SurroundOcc~\citep{yiwei2023surroundocc} performs 3D BEV feature queries from multi-view, multi-scale 2D image features, incorporates 3D convolutions into Transformer layers, and progressively upsamples volume features. The hierarchical fusion strategy enables dense occupancy prediction while preserving geometric details in complex urban scenes. However, most methods limit their representation to current 3D space without considering future states of surrounding objects over time. To extend camera-only occupancy estimation to spatiotemporal prediction, Cam4DOcc~\citep{ma2024cam4docc} introduces a new benchmark that extends occupancy prediction by estimating both current and future environmental states. Cam4DOcc proposes an end-to-end 4D occupancy forecasting network, using multi-frame feature aggregation to extract 3D voxel features from sequential camera images and a future state prediction module to forecast occupancy.

In the 3D object tracking task, most existing methods rely on heuristic strategies that use spatial and appearance similarities. However, they often fail to effectively model temporal information. Recent Transformer-based models address these limitations through novel query mechanisms. For example, MOTR \citep{motr2022zeng} extends the DETR model and builds a multiple-object tracking (MOT) framework. It introduces a "track query" to model tracked instances throughout a video, designed to exploit temporal variations in video sequences and implicitly learn long-term temporal changes in targets, avoiding the need for explicit heuristic strategies. Unlike traditional methods that rely on motion-based and appearance-based similarity heuristics and post-processing techniques, MOTR handles object tracking without requiring track NMS or IoU matching. Building on this foundation, MUTR3D~\citep{zhang2022mutr3d} simultaneously performs detection and tracking by employing cross-camera and cross-frame object association based on spatial and appearance similarities. This approach utilizes a 3D track query to directly model an object's 3D states and appearance features over time and across multiple cameras. During each frame, the 3D track query samples features from all visible cameras and learns to initiate, track, or terminate tracks. MeMOTR~\citep{gao2023memotr} further advances the field by introducing a long-term memory-augmented Transformer for multi-object tracking. MeMOTR uses track queries but enhances them by incorporating a long-term memory mechanism. It injects longer temporal information into the tracking process, significantly improving the performance of multi-target tracking. However, query-memory designs still rely heavily on current-frame association cues in difficult occlusion cases. TrajectoryFormer~\citep{chen2023trajectoryformer} complements these approaches by introducing predictive trajectory hypotheses into the Transformer tracking pipeline, so that future motion priors can directly guide association and state update. This design is particularly useful in crowded and short-term occlusion scenarios, where pure appearance matching is often ambiguous. However, despite these advancements in temporal information modeling, most of these methods focus primarily on the data association task while neglecting the accurate estimation of object states such as velocity and acceleration. STT~\citep{jing2024stt} addresses this gap through a unified joint optimization framework. It tracks objects by leveraging rich appearance, geometry, and motion signals from long-term detection history and jointly optimizes both tasks. STT emphasizes the importance of joint estimation tasks for autonomous driving, and to address the limitations of existing evaluation methods, it extends the MOTA metric to S-MOTA and extends the MOTP metric to MOTPs, enforcing consideration of state estimation accuracy in tracking evaluation, thereby capturing more accurate motion states of objects.

These developments highlight the growing potential of Transformer-based methods in complex 3D perception tasks. Current innovation focuses on specialized attention mechanisms for spatiotemporal fusion, but future work must address the challenges of multi-object scenarios in dense urban environments.

\subsection{2D \& Plane Tasks}
In contrast to the 3D task category, we classify a second task category as 2D and plane tasks where models primarily address tasks such as lane detection, segmentation, and HD map generation. 

For the lane detection task, we divide the models into two groups. The first group of models generate BEV features followed by CNN semantic segmentation and detection heads. For example, BEVSegFormer~\citep{peng2023bevsegformer} uses a cross-attention mechanism to query multi-view 2D image features, with a semantic decoder converting query outputs into BEV road segmentation results. PersFormer~\citep{persformer2022lichen} extracts image features using a CNN and splits them into two paths. The first path connects to a CNN-based 2D lane detection head, while the second path uses an inverse perspective mapping (IPM) method to transform perspective view (PV) view features to BEV view features, connected to the Transformer network for BEV feature querying and enhancement. The second group queries and generates road structures using representations such as polynomials, key points, vectors, and polylines. For instance, LSTR~\citep{lstr2021ruijin} approximates flat single-lane road markings with second- or third-order polynomials. A Transformer query updates the polynomial parameters, and the Hungarian matching loss optimizes the path-related regression loss. LSTR adopts a lightweight Transformer architecture for efficient querying. CurveFormer~\citep{bai2022curveformer} accelerates inference by generating lane lines directly from 2D images without feature view transformations. It employs a Transformer decoder using curve queries to transform the 3D lane detection formula into a curve propagation problem, with a curve intersection attention module calculating similarity between curve queries and image features. Extending curve-query paradigms, CurveFormer++~\citep{bai2025curveformer++} introduces dynamic anchor sets, curve cross-attention, and temporal query propagation, demonstrating that sparse lane-centric temporal modeling can improve both 3D lane detection accuracy and frame-level stability. LATR~\citep{luo2023latr} introduces a lane-aware query generator constructing queries using hybrid embeddings that combine lane-level and point-level features. Additionally, it employs a dynamic 3D ground positional embedding injecting 3D spatial information into 2D image features, enabling the model to capture the 3D structure of lanes accurately. This approach avoids misalignment issues in methods relying on BEV transformations.

However, these methods still expose two practical gaps: robustness under severe visual degradation and temporal consistency across consecutive frames. More recent works explicitly target these issues. HWLane~\citep{zhao2024hwlane} proposes row-column constrained self-attention (HW-Transformer) and self-attention knowledge distillation, achieving strong results under night, glare, and occlusion conditions while retaining efficient inference. 

Beyond lane detection, Transformer architecture is also utilized in segmentation tasks. For example, TIiM~\citep{tiim2022saha} presents a sequence-to-sequence model for instantaneous mapping that converts images and videos into overhead maps or BEV representations. By assuming a one-to-one correspondence between vertical scan lines in images and rays in overhead maps, TIiM is a data-efficient and spatially-aware method. Panoptic SegFormer~\citep{panoptic2022li} proposes a framework for panoptic segmentation combining semantic and instance segmentation. It introduces a supervised mask decoder and query decoupling strategy to perform efficient segmentation.  

For HD map generation, STSU~\citep{stsu2021yigit} represents lanes as directed graphs in BEV coordinates and learns Bezier control points and graph connectivity through multilayer perceptrons (MLPs). It adopts a DETR-type query method to convert front-view camera images into BEV road structures. VectorMapNet~\citep{liu2023vectormapnet} extends this paradigm as the first Transformer network achieving end-to-end vectorization for high-precision maps~\citep{qili2021hdmapnet}. It models geometric shapes using sparse polyline primitives from BEV view, implementing a two-stage pipeline consisting of set prediction for detecting rough key points and sequence generation for predicting map elements' next points. MapTR~\citep{maptr2023bencheng} develops a framework for online vectorized high-precision map generation modeling map elements as point sets with equivalent envelopes. It introduces hierarchical query embeddings to flexibly encode instance-level and point-level information and learns structured bipartite matching of map elements. However, VectorMapNet and MapTR face limitations in perception range and temporal information. They are typically constrained to small perception ranges and rely solely on single-frame inputs, making them susceptible to errors in complex driving scenarios. To address these limitations, StreamMapNet~\citep{yuan2024streammapnet} proposes a novel streaming mapping pipeline incorporating multi-point attention and temporal information. This approach employs a streaming strategy for temporal fusion, propagating hidden states across frames to preserve historical information while minimizing memory/latency costs. It expands the perception range to 100 × 50 meters and improves temporal consistency, significantly enhancing map construction stability and performance. These models effectively merge multi-view features into a unified BEV view, facilitating end-to-end online high-precision map construction crucial for downstream tasks. Recent studies further strengthen online vectorized mapping from both architectural and temporal perspectives. CAFMap~\citep{fang2025cafmap} explores an end-to-end convolution-attention fusion design for vectorized HD map construction, highlighting the importance of combining local geometric encoding with global relational reasoning. Building on this temporal direction, PrevPredMap~\citep{peng2025prevpredmap} explicitly injects previous predictions into the current-frame mapping process, showing that prediction-level temporal priors can improve online map consistency beyond conventional feature-only temporal fusion.

\subsection{Prediction and Decision Making}

Following perception-centric tasks, autonomous driving systems must further address downstream trajectory prediction, planning, and decision-making under closed-loop constraints. Transformer architectures are therefore increasingly adopted in these modules, and recent studies further explore end-to-end deep neural network (DNN) designs that unify perception, prediction, and control into a single pipeline.

For trajectory or behavior prediction, practical challenges exist in feature extraction with standard CNN models, especially regarding their limited capacity to model long-range interactions. Transformer-based models are then developed to address this issue. VectorNet~\citep{vectornet2020gao} is developed to transform these geometric shapes (from road markings or vehicle trajectories) into vector format inputs. It introduces a hierarchical graph neural network encoding HD maps and agent trajectories using vector representations, exploiting spatial locality of individual road components and modeling their interactions. TNT~\citep{tnt2021zhao} defines vehicle modes based on the endpoint of each trajectory and simplifies trajectory prediction by converting it into an endpoint prediction problem. However, as an anchor-based technique, TNT requires heuristic anchor definitions before predicting the endpoint. DenseTNT~\citep{densetnt2021gu} overcomes this limitation by directly predicting the probability distribution of the endpoint, enabling anchor-free prediction. mmTransformer~\citep{mmTransformer2022liu} proposes a stacked Transformer architecture to model multi-modality at the feature level with a set of fixed independent proposals. A region-based training strategy is then developed to induce the multi-modality of the generated proposals. This strategy reduced complexity of motion prediction while ensuring multimodal behavior outputs. AgentFormer~\citep{agentformer2021yuan} allows an agent's state at a specific time to directly impact another agent's future state, eliminating the need for intermediate features encoded in a single dimension. This approach enables the simultaneous learning of temporal information and interaction relationships. It preserves temporal coherence while addressing information loss in traditional attention mechanisms caused by equal input element status. For complex situations with static and dynamic inputs (e.g., road geometry, lane connectivity, traffic lights, etc.), standard Transformers struggle to model extensive multi-dimensional sequences due to self-attention's quadratic computational complexity relative to input sequence length and costly position-wise feed-forward networks. WayFormer~\citep{wayformer2022nayakanti} mitigates this by analyzing pre-fusion, post-fusion, and hierarchical fusion strategies for inputs, maintaining efficiency-quality balance. This method avoids designing modality-specific modules, facilitating model scaling and extension. HPTR~\citep{zhang2023real} advances the field using K-nearest Neighbor Attention with Relative Pose Encoding (KNARPE), which encodes relative position and attitude information between agents. It minimizes computational overhead through context sharing among agents and asynchronous token updates during online inference. Additionally, the Co-MTP framework~\citep{zhang2025co} introduces a cooperative trajectory prediction approach with multi-temporal fusion for autonomous driving. This framework uses V2X technology to capture interactions among agents in historical and future domains, significantly improving prediction accuracy. To more directly link prediction quality to safety, IA-TTLP~\citep{liang2025interaction} proposes a Transformer-transfer learning framework that first learns from large-scale HDV data and then adapts to limited AV-HDV interaction data, while explicitly modeling prediction uncertainty for downstream planning constraints.

Finally, end-to-end integrated driving frameworks aim to produce planning or control outputs directly from multimodal sensory observations. Several works have emerged in the past few years. For example, TransFuser~\citep{transfuser2022chitta} uses multiple Transformer modules for data processing, intermediate data fusion, and feature map generation. Data fusion is applied at multiple resolutions (64×64, 32×32, 16×16, and 8×8) throughout the feature extractor, resulting in a 512-dimensional feature vector output from both the image and LiDAR BEV streams, which are combined via element-wise summation. The approach considers sensing regions within 32m in front of the ego-vehicle and 16m to each side, thereby encompassing a BEV grid of 32m × 32m. This grid is divided into blocks of 0.125m × 0.125m, resulting in a resolution of 256 × 256 pixels. NEAT~\citep{neat2021chitta} proposes a representation for efficient reasoning of semantic, spatial, and temporal structure of scenes. It constructs a continuous function that maps locations in BEV scene coordinates to waypoints and semantics, using intermediate attention maps to iteratively compress high-dimensional 2D image features into a compact representation. Building on the TransFuser architecture, InterFuser~\citep{interfuser2023shao-safety} proposes a one-stage architecture fusing multi-modal, multi-view sensor information, achieving enhanced performance. This framework enhances the safety of the end-to-end model by developing a safety control filter to constrain Transformer output actions. The model's safety-insensitive outputs consist of a 10-waypoint path, while the safety-sensitive outputs include traffic rule information and an object density map with seven features for objects such as vehicles, pedestrians, and bicycles. These outputs are produced by fusing multi-view image inputs and LiDAR point cloud data, which cover a region extending 28 meters in front of the ego-vehicle and 14 meters to its sides. The analyzed area measures 20m × 20m and is divided into 1m × 1m grids. In addition to camera and LiDAR signals, MMFN~\citep{zhang2022mmfn} makes use of vectorized HD maps and radar in the end-to-end task. It explores different representations of the HD map as the network input and proposes a framework fusing all four data types. STP3~\citep{stp32022hu} proposes an egocentric aligned accumulation scheme converting 2D to 3D while aligning the target features. Its prediction module integrates information from both the obstacle at time $t$ and the obstacle position at time $t-n$. ReasonNet~\citep{shao2023reasonnet} is a novel end-to-end driving framework. It leverages both temporal and global reasoning to enhance perception and decision-making. The framework includes a temporal reasoning module to process historical scene information and a global reasoning module to model interactions among objects and the environment, improving overall perception performance and safety. Scene-Rep Transformer~\citep{liu2024augmenting} enhances RL decision-making capabilities through improved scene representation encoding and sequential predictive latent distillation. It uses a multi-stage Transformer to model interactions between the ego vehicle and its neighbors, and a sequential latent Transformer to distill future predictive information into the latent scene representation, enabling better decision-making and more diversified driving behaviors.

Alongside these direct end-to-end pipelines, planning-oriented unified architectures place planning at the center of system design. Unlike the works mentioned above that are primarily designed as end-to-end driving pipelines, UniAD~\citep{uniad2023hu} presents a planning-oriented framework. The paper argues that previous works failed to consider certain components necessary for planning, and the new design could properly organize preceding tasks to facilitate planning. Recently, BridgeAD~\citep{zhang2025bridging} introduces a novel framework that enhances autonomous driving by integrating historical prediction and planning information across perception, prediction, and planning stages. The core idea of BridgeAD is to represent motion and planning queries as multi-step queries, differentiating each future time step to effectively leverage historical insights, thereby improving the accuracy and coherence of perception and motion planning. In the significant shift toward more efficient and diverse planning and decision-making, DiffusionDrive~\citep{liao2025diffusiondrive} introduces a novel truncated diffusion model. By incorporating prior multi-mode anchors and a truncated diffusion schedule, the model learns denoising from anchored Gaussian distributions to generate multi-mode driving actions. This approach delivers superior diversity and quality in just 2 steps, addressing the limitations of conventional methods and setting a new benchmark in the field.

Recent advancements further extend this direction toward language-augmented decision making. The success of GPT has inspired autonomous driving researchers to integrate large language models (LLMs) and vision-language models (VLMs) into driving frameworks. These models improve semantic reasoning and human-vehicle interaction by combining multimodal perception with language understanding. LLMs can serve as externalized knowledge repositories, provide instruction-following behavior, and improve few-shot generalization in long-tail scenarios. DriveGPT4~\citep{xu2024drivegpt4} is an early interpretable system in this line, producing low-level driving actions together with natural-language rationales from multi-frame visual inputs and textual queries. Furthermore, several studies combine language models with vision-centric autonomous driving pipelines. VLP~\citep{pan2024vlp} bridges language understanding and autonomous driving by generating enhanced BEV features for planning-oriented inference. DriveVLM~\citep{tian2024drivevlm} employs VLM reasoning and chain-of-thought style decomposition for hierarchical planning. Unlike purely data-driven policies, DiLU~\citep{wen2023dilu} introduces a knowledge-driven framework that accumulates driving experience via memory, reasoning, and reflection modules. Building on these ideas, SimLingo~\citep{renz2025simlingo} proposes a unified Vision-Language-Action framework for closed-loop driving, vision-language understanding, and language-action alignment.

Overall, Transformer-based research in prediction and decision making is evolving from single-task predictors to end-to-end integrated driving frameworks, then toward planning-oriented unified architectures and language-augmented decision-making systems. This evolution improves capability and interpretability, but also amplifies deployment pressure due to higher latency, memory footprint, and compute demand. Therefore, after reviewing the task-level model landscape, the following sections focus on benchmarked runtime-performance trade-offs and the necessity of model compression for real-world deployment.

\begin{table}[!ht]
\centering
\caption{\label{tab:nVidiaBenchmark}Benchmark performance of Transformer models on a standard NVIDIA RTX 3090 GPU. `Backbone' describes the backbone architecture of each model and `Param.' represents the size of model parameters. `GFLOPS' stands for giga floating point operations per second, `FPS' is frames per second, and `mAP' represents mean average precision of the model. The last column indicates the benchmark dataset used for each model.}
\scriptsize
\setlength{\tabcolsep}{3pt}
\begin{adjustbox}{max width=\textwidth}
\begin{tabular}{@{}ccccccccc@{}}
\toprule
\textbf{Task} & \textbf{Model} & \textbf{Backbone} & \textbf{Input Size} & \textbf{Param.} & \textbf{GFLOPs} & \textbf{FPS} & \textbf{mAP} & \textbf{Dataset} \\
\midrule
\multirow{2}*{Base Model} & ViT~\citep{dosovitskiy2020image} & - & 224x224 & 85.8M & 16.86 & 14.0 & 0.81 & ImageNet~\citep{deng2009imagenet}\\
~ & Swin-T~\citep{liu2021swin} & - & 224x224 & 86.8M & 15.14 & 27 & 0.81 &ImageNet\\
\midrule
\multirow{13}*{Object Detection} & DETR~\citep{carion2020end} & ResNet50 & 800x800 & 42.22M & 73.4 & 9 & 0.385 &Coco~\citep{lin2014microsoft}\\
~ & DETR3D~\citep{wang2022detr3d} & ResNet101 & 6x1600x900 & 51.3M & 1016.8 & 2.0 & 0.349 &nuScenes~\citep{caesar2020nuscenes}\\
~ & FUTR3D~\citep{chen2023futr3d} & ResNet101  & 6x1600x900 & 47.7M  & 1023 & 1.9 & 0.346 &nuScenes\\
~ & BEVFormer~\citep{li2022bevformer} & ResNet101 & 6x1600x900 & 68.7M & 1303.5 & 2.3 & 0.375 &nuScenes\\
~ & PETR~\citep{liu2022petr} & ResNet101 & 6x1408x512 & 59.2M & 504.6 & 5.3 & 0.357 &nuScenes \\
~ & PETR & ResNet50 & 6x1408x512 & 35.8M & 290.0 & 10.4 & 0.341 &nuScenes\\
~ & PETR & VoVNet & 6x640x1600 & 81.66M & - & 13.4 & 0.40 &nuScenes\\
~ & StreamPETR~\citep{wang2023exploring} & ResNet50 & 6x256x704 & 37.21M & - & 21.5 & 0.42 &nuScenes\\
~ & StreamPETR & VoVNet & 6x320x800 & 82.95M & - & 10.3 & 0.48 &nuScenes\\
~ & CrossDTR~\citep{tseng2022crossdtr} & ResNet101 & 6x1408x512 & 53.3M & 483.9 & 5.8 & 0.370 &nuScenes\\
~ & CrossDTR & ResNet50 & 6x1408x512 & 31.8M & 268.1 & 10.6 & 0.326 &nuScenes\\
~ & \multirow{2}*{IS-Fusion~\citep{yin2024fusion}}& \multirow{2}*{Swin-T} & 1440×1440×40(lidar) & \multirow{2}*{48.32M} & \multirow{2}*{-} & \multirow{2}*{2.7} & \multirow{2}*{0.73} & \multirow{2}*{nuScenes} \\
~ & ~ & ~ & 6×384×1056(camera) & ~ & ~ & ~ & ~ & ~\\
\midrule
\multirow{3}*{Lane Detection} & LSTR~\citep{lstr2021ruijin}  & ResNet18 & 1x640x360 & 765.8K & 1.15 & 89.6 & 0.962 & TuSimple~\citep{yoo2020end}\\
~ & Persformer~\citep{persformer2022lichen} & EfficientNet & 1x480x360 & 47.7M & 142.8 & 24.03 & 0.505 & OpenLane~\citep{chen2022persformer}\\
~ & LATR~\citep{luo2023latr} & ResNet50 & 1x720x960 & 46.8M &  & 38.01 & 0.41 & OpenLane\\
\midrule
\multirow{2}*{HD Map} & MapTR~\citep{maptr2023bencheng} & ResNet50 & 6x800x450 & 36.2M & - & 16.6 & 0.762 &nuScenes\\
~ & MapTR & ResNet18 & 6x320x180 & 15.4M & - & 34 & 0.642 &nuScenes\\
\midrule
\multirow{2}*{Prediction} & DenseTNT~\citep{densetnt2021gu} & - & 1x296x128 & 617.8K & 39.23 & 7.41 & 0.328 &Argoverse~\citep{chang2019argoverse} \\
~ & HPTR~\citep{zhang2023real} & - & - & 15.2M & - & 24.1 & 0.30 &Argoverse\\
% Agentformer & Designed CNN  &  &  &  &  &  & ETH, UCY, nuScenes \\
% WayFormer &  &  & 0.3M-20M &  & 8-64 &  & Waymo \\
\midrule
\multirow{4}*{End-to-End} & \multirow{2}*{Transfuser~\citep{transfuser2022chitta}} & RegNetY & 256x256 (LiDAR) & \multirow{2}*{165.6M} & \multirow{2}*{33.8} & \multirow{2}*{59.6} & \multirow{2}*{61.18} & \multirow{2}*{CARLA~\citep{dosovitskiy2017carla}} \\
~ & ~ & RegNetY & 704x160 (RGB) & ~ & ~ & ~ & ~ & ~ \\
~ & \multirow{2}*{Interfuser~\citep{interfuser2023shao-safety}}& ResNet18 & 224x224 (LiDAR) & \multirow{2}*{82.8M} & \multirow{2}*{46.5} & \multirow{2}*{38.2} & \multirow{2}*{71.18} & \multirow{2}*{CARLA} \\
~ & ~ & ResNet50 & 800x600 (RGB) & ~ & ~ & ~ & ~ & ~\\
\bottomrule
\end{tabular}
\end{adjustbox}
\end{table}

\subsection{Benchmark of Transformer Models}

Table~\ref{tab:nVidiaBenchmark} summarizes representative Transformer-based models using commonly reported indicators such as input size, runtime, accuracy, and evaluation dataset, with supplementary measurements conducted on an NVIDIA RTX 3090 where feasible. These results should be interpreted as indicative rather than strictly apples-to-apples, because the models differ in task formulation, sensor configuration, backbone, preprocessing, and evaluation protocol. Within the 3D object detection setting on nuScenes~\citep{caesar2020nuscenes}, DETR3D and FUTR3D exhibit broadly similar performance owing to their related query-based formulations. BEVFormer improves over DETR3D by constructing BEV representations and querying 3D objects from those features. PETR and CrossDTR further transform 2D image features into 3D-aware representations using CNN-based modules, which accelerates query processing and improves performance relative to DETR3D. ResNet101-based variants generally deliver higher accuracy than ResNet50-based variants, likely due to increased backbone capacity and, in some cases, the use of deformable convolutions, although this usually comes at the expense of runtime~\citep{jifeng2017dcn}. In contrast, Transformer-based road-element detection shows greater heterogeneity, with different model assumptions and evaluation protocols across 2D lane detection (TuSimple), 3D lane detection (OpenLane), and local-map construction (nuScenes). Lane and local-map models are typically faster than 3D object detectors because they require fewer key-point queries and often use lighter CNN backbones with shallower feature hierarchies. As shown in the lower portion of the table, end-to-end Transformer driving remains an emerging direction. However, many such systems still rely heavily on simulated platforms such as CARLA~\citep{dosovitskiy2017carla}, which limits direct conclusions about real-world deployment readiness.

\subsection{Model Compression Imperative}
\label{sec:model_compression}
In autonomous driving, Transformer architectures are steadily evolving toward larger multimodal and multitask systems with increasingly rich spatiotemporal inputs and outputs. This trend raises the demands placed on training, inference, and hardware support. The inherent complexity of Transformer models often leads to substantial computational overhead and memory consumption, making real-time deployment difficult without dedicated optimization of both representation learning and the encoder--decoder pipeline. As Table~\ref{tab:nVidiaBenchmark} illustrates, architectural design alone can improve the efficiency--accuracy trade-off to a meaningful extent. For example, CrossDTR (ResNet101) reduces parameters to 53.3M through lightweight design---a 22\% reduction relative to BEVFormer (68.7M) for the same task---while maintaining comparable accuracy and increasing throughput from 2.3 to 5.8 FPS. Similarly, StreamPETR (VoVNet) sustains 0.48 mAP at 10.3 FPS through more efficient temporal modeling.
However, software-level redesign alone is usually insufficient to satisfy the stringent real-time requirements of onboard deployment. Such optimizations can reduce computational overhead and improve inference speed, but they often introduce accuracy trade-offs or leave substantial hardware inefficiencies unresolved. For example, replacing ResNet101 with ResNet50 in CrossDTR increases throughput from 5.8 to 10.6 FPS but reduces mAP by 11.9\% (from 0.37 to 0.326).
Consequently, model compression has become indispensable for balancing efficiency, robustness, and deployability. Techniques such as quantization, pruning, knowledge distillation, and optimized attention mechanisms can substantially reduce computational cost and memory footprint, thereby helping Transformer models satisfy the latency and efficiency constraints of practical autonomous driving systems.

\section{Model Compression for Transformer-based Autonomous Driving Models}
\label{Sec: compression}
As discussed in Section~\ref{sec:model_compression}, the limitations of software-level optimizations motivate a more fundamental approach. Rather than deploying small full-precision models, compression techniques aim to preserve the emergent capabilities of large-scale Transformer models while systematically eliminating computational and parametric redundancies. Through quantization, pruning, knowledge distillation, and low-rank decomposition, this approach achieves a significantly more favorable trade-off between system efficiency and task-level robustness.

Common compression strategies include quantization, pruning, knowledge distillation, and low-rank decomposition, each targeting different sources of redundancy within Transformer networks. Quantization mitigates memory bottlenecks by converting high-precision floating-point parameters into compact low-bit representations. This reduction in bandwidth demand is crucial for accelerating matrix operations on automotive-grade hardware. In practice, quantization can be implemented either as post-training quantization (PTQ), which directly calibrates a trained model for low-bit inference, or as quantization-aware training (QAT), which introduces quantization effects during training to improve robustness. For Transformer-based autonomous driving models, this process is not merely a numerical compression step, because attention logits, softmax activations, and positional encoding branches are often more sensitive to dynamic-range mismatch than standard convolutional layers. Binarization pushes this concept to its limit by constraining values to binary states, significantly reducing arithmetic complexity for emerging low-power accelerators. Pruning removes redundant weights, channels, or attention heads to streamline computation, which is particularly beneficial for multi-view, multi-sensor perception models. Knowledge distillation transfers representational and task-level knowledge from a large teacher model to a compact student model, helping maintain robustness in long-tail or safety-critical scenarios. In practical systems, distillation is usually realized through output-logit supervision, intermediate feature alignment, or relation-level consistency constraints, so that a lightweight student can preserve the teacher's task-specific behavior. This is especially important in autonomous driving, where model compression must retain geometric reasoning, cross-modal alignment, and safety-relevant predictions rather than only reducing parameter count. Low-rank decomposition factorizes large projection matrices within self-attention or feed-forward layers, lowering both storage and multiply–accumulate operations without modifying the overall architecture. As compression algorithms continue to advance, these techniques play a central role in alleviating the computational bottlenecks of real-time autonomous driving, enabling large-scale Transformers to be deployed efficiently on resource-limited onboard platforms.

\begin{table}[!ht]
\centering
\caption{\label{tab:Compression-modelTable} Representative compression methods reported for Transformer-based autonomous driving models, grouped by task domain and compression strategy.} 

\small
\setlength{\tabcolsep}{5pt}
\begin{adjustbox}{max width=\textwidth}
\begin{tabular}{@{}llll@{}}

\toprule
\textbf{Model} & \textbf{Tasks} & \textbf{Compression} & \textbf{Release Year} \\
\midrule
DetPTQ~\citep{niu2023improving} & Object Detection & Post-Training Quantization & 2023 \\
LiDAR-PTQ~\citep{zhou2024lidar} & Object Detection & Post-Training Quantization & 2024 \\
Q-PETR~\citep{yu2025q} & Object Detection & Post-Training Quantization & 2025 \\
Multi-Modal Joint Input Pruning~\citep{li2024learning} & Object Detection & Pruning & 2024 \\
PointDistiller~\citep{zhang2023pointdistiller} & Object Detection & Knowledge Distillation & 2022 \\
BEVDistill~\citep{chen2022bevdistill} & Object Detection & Knowledge Distillation & 2022 \\
DistillBEV~\citep{wang2023distillbev} & Object Detection & Knowledge Distillation & 2023 \\
LiDAR2Map~\citep{wang2023lidar2map} & HD Map Generation & Knowledge Distillation & 2023 \\
MapDistill~\citep{hao2024mapdistill} & HD Map Generation & Knowledge Distillation & 2024 \\
MapKD~\citep{yan2025mapkd} & HD Map Generation & Knowledge Distillation & 2025 \\
PlanKD~\citep{feng2024road} & End-to-End & Knowledge Distillation & 2024 \\
OWLed~\citep{li2024owled} & End-to-End & Pruning & 2024 \\
FastDriveVLA~\citep{cao2025fastdrivevla} & End-to-End & Pruning & 2025 \\
DSDrive~\citep{liu2025dsdrive} & Large Language Models & Pruning & 2025 \\

\bottomrule
\end{tabular}
\end{adjustbox}
\end{table}

\subsection{Task-specific Compression Methods}

\subsubsection{3D Object Detection and Segmentation}
Transformer models for 3D object detection face a dual challenge: they must process high-dimensional data, such as point clouds or multi-view images, while maintaining precise localization. In this context, model quantization is often prone to accuracy degradation, as different layers exhibit varying sensitivity to quantization noise. Compared with conventional 2D vision models, 3D detection pipelines are usually more vulnerable to low-bit perturbation because geometric projection, sparse sampling, and cross-view attention all amplify numerical error in downstream localization. Therefore, effective quantization methods in this domain often require task-aware calibration rather than uniform low-bit conversion across all operators. DetPTQ~\citep{niu2023improving} is a representative solution to this challenge. It uses the Object Detection Output Loss (ODOL) to assign different quantization parameters to different layers. This design effectively reduces computational complexity while maintaining detection accuracy, and alleviates the performance degradation problem of traditional PTQ (Post-Training Quantization) in complex perception tasks.

Specific challenges arise when dealing with irregular data modalities. Unlike 2D images, point clouds exhibit strong sparsity and irregular geometry, making PTQ particularly unstable. Although PTQ has been widely applied in 2D computer vision tasks, directly transferring PTQ methods from the image domain to 3D point cloud tasks leads to severe performance degradation. Therefore, considering the uniqueness of the point cloud modality, LiDAR-PTQ~\citep{zhou2024lidar} is proposed. LiDAR-PTQ innovatively presents three key components: Sparsity-based Calibration to initialize quantization parameters; Task-guided Global Positive Loss (TGPL) to reduce the disparity between the final predictions before and after quantization; adaptive rounding-to-nearest operation to minimize the layerwise reconstruction error. In LiDAR-based 3D detection tasks, LiDAR-PTQ enables the PTQ INT8 model to achieve accuracy almost identical to the FP32 model while enjoying 3× inference speedup. In addition to the general challenges of quantizing Transformer-based 3D detection models, for multi-view camera-based methods like the PETR series, a core difficulty lies in the extreme mismatch between the dynamic ranges of positional encodings and image features. The PETR series is a mainstream Transformer-based framework for multi-view 3D object detection, known for its excellent performance. However, direct INT8 quantization leads to substantial accuracy degradation. To address this issue, the Q-PETR~\citep{yu2025q} model re-engineers key components of the PETR framework to reconcile this discrepancy and to adapt the cross-attention mechanism for low-bit inference. Specifically, Q-PETR redesigns the positional encoding module and introduces an adaptive quantization strategy. It adopts amplitude-aligned positional encoding, performs softmax quantization only after numerical stability is achieved, and designs a dual-level lookup table (duLUT) for non-linear operators. For the first time, the PETR series can be deployed under standard 8-bit post-training quantization with virtually no accuracy loss.

Quantization focuses on reducing numerical precision, while knowledge distillation (KD) provides a complementary approach by transferring the capabilities of a large model to a compact one. It works by transferring knowledge from an over-parameterized trained model (teacher model) to a lightweight student model, and has demonstrated consistent effectiveness in 2D vision tasks. However, in autonomous driving, the purpose of distillation is not only to preserve classification accuracy but also to retain geometric structure understanding, BEV representation quality, and cross-modal consistency after model lightweighting. As a result, effective distillation usually goes beyond final-logit supervision and incorporates feature-level or relation-level guidance. However, due to the inherent sparsity and irregularity of point cloud data, directly applying traditional image-based knowledge distillation methods to point cloud detection often fails to achieve desired results. To fill this gap, PointDistiller~\citep{zhang2023pointdistiller} is the first to propose a structured knowledge distillation framework specifically designed for point cloud-based 3D detection. It introduces a local distillation mechanism that extracts and distills knowledge about the local geometric structure of point clouds through dynamic graph convolution, enabling the student model to better understand 3D spatial structures. Simultaneously, it adopts a reweighted learning strategy that dynamically adjusts the distillation weights based on the number of points contained within each voxel, emphasizing the learning of key regions to address the sparsity and noise issues of point clouds. Building upon the same-modal distillation framework established by PointDistiller, BEVDistill~\citep{chen2022bevdistill} extends this paradigm to a more challenging and practically relevant cross-modal setting, enabling effective knowledge transfer from LiDAR to camera-based models. BEVDistill aligns heterogeneous features in a unified bird’s-eye-view (BEV) representation and distills knowledge from a high-capacity LiDAR-based teacher to an image-only student. This transfer is realized through two complementary mechanisms: dense feature distillation at the BEV level and sparse instance-level distillation, jointly facilitating robust cross-modal representation learning. Uniform distillation overlooks the uneven spatial distribution of semantic cues in BEV space. Building upon this foundation, DistillBEV~\citep{wang2023distillbev} proposes a more refined and comprehensive distillation framework. It introduces region decomposition to enable differentiated distillation, adaptive scaling to balance learning weights for objects of different sizes, and a spatial attention mechanism that allows the student model to focus on features deemed critical by the teacher model. Furthermore, DistillBEV extends distillation to multi-scale layers and temporal fusion, achieving a more comprehensive feature alignment. This systematic strategy enables DistillBEV to achieve significant performance improvements in multiple representative student models (such as the BEVDet series, BEVDepth, and BEVFormer).

\subsubsection{HD Map Construction and Lane Detection}
Although dynamic object detection has been a major focus in autonomous driving, understanding static scenes, particularly high-definition (HD) map construction, introduces distinct challenges. Unlike detection tasks, HD mapping requires both precise geometric reconstruction and rich semantic understanding, so compression must preserve structural completeness and topological consistency in addition to improving efficiency. Camera-based methods offer rich semantics but poor geometry, while LiDAR provides precision but lacks texture. Fusion models bridge this gap but incur high costs, necessitating efficient compression strategies that can leverage multi-modal strengths without the deployment burden. To address these issues, LiDAR2Map~\citep{wang2023lidar2map} is the first to verify the feasibility of Camera-to-LiDAR distillation. Aiming at the pain points of LiDAR-BEV, such as weak texture and heavy noise, it designs a lightweight BEV Feature Pyramid Decoder (BEV-FPD), which acquires clean and sharply responsive bird's-eye view features through 6-level residual multi-scale aggregation. To mitigate the defects caused by lacking semantic cues in LiDAR data, they present an online Camera-to-LiDAR distillation scheme to facilitate the semantic learning from image to point cloud. The distillation scheme consists of feature-level and logit-level distillation to absorb the semantic information from camera in BEV, enabling effective semantic alignment in the BEV space. MapDistill~\citep{hao2024mapdistill} is a novel knowledge distillation framework designed to improve the efficiency and accuracy of camera-only high-definition (HD) map construction by distilling knowledge from camera–LiDAR fusion models. To enable effective cross modal knowledge transfer, MapDistill introduces a dual BEV transform module that aligns representations across modalities in the bird's-eye view space. Furthermore, MapDistill proposes a comprehensive distillation scheme, which includes cross-modal relation distillation, dual-level feature distillation, and map head distillation. Specifically, cross-modal relation distillation enables the student model to learn the cross-modal representations of the teacher model by aligning attention matrices; dual-level feature distillation aligns low-level and high-level features in the BEV space; and map head distillation uses the predictions of the teacher model as pseudo-labels to supervise the student model's classification and point regression tasks. This method achieves a balance between cost and accuracy, providing a strong baseline for efficient HD map construction. Further advancing this direction, MapKD~\citep{yan2025mapkd} introduces a novel Teacher-Coach-Student (TCS) distillation framework that bridges the cross-modal knowledge transfer gap by incorporating a coach model. Furthermore, MapKD designs two targeted distillation strategies: Token-Guided 2D Patch Distillation (TGPD) aligns BEV features via patch-token attention to enhance spatial detail learning, while Masked Semantic Response Distillation (MSRD) optimizes semantic prediction by focusing on foreground regions, thereby improving semantic consistency. MapKD performs well both quantitatively and qualitatively, offering a flexible and efficient solution without relying on HD map pretraining.

\subsubsection{End-to-End Driving Systems}
The methods discussed above primarily focus on individual perception modules. However, the industry is increasingly shifting towards End-to-End Driving Systems that integrate perception, prediction, and planning into a unified network. While promising, the massive computational overhead of these monolithic networks limits their deployment on resource-constrained devices, making compression essential for practical adoption. Because these models are ultimately evaluated by closed-loop driving quality rather than isolated perception metrics, compression must be considered together with planning reliability and safety-oriented performance. To address this issue, knowledge distillation has been introduced for model compression, and PlanKD~\citep{feng2024road} is the first knowledge distillation framework tailored specifically for compressing end-to-end motion planners. Firstly, considering that driving scenarios are inherently complex and usually contain information irrelevant to planning or even noisy data, transferring such information is of no benefit to the student model. Therefore, a strategy based on the information bottleneck is designed to distill only the information relevant to planning, rather than transferring all information indiscriminately. Secondly, different waypoints in the output planned trajectory may have varying degrees of importance for motion planning—minor deviations in certain key waypoints can even lead to collisions. Thus, a safety-aware waypoint-focused distillation module is designed, which assigns adaptive weights to different waypoints according to their importance. This encourages the student model to accurately mimic the more critical waypoints, thereby improving overall safety.

Unlike obtaining lightweight models using distillation methods, FastDriveVLA~\citep{cao2025fastdrivevla} accelerates the inference process through pruning. To address the issue of visual token redundancy in Vision-Language-Action (VLA) models, FastDriveVLA proposes an innovative token pruning framework. It focuses on directly reducing the number of visual tokens and achieves plug-and-play pruning via its core component, ReconPruner. ReconPruner adopts MAE-style pixel reconstruction training and combines it with an adversarial foreground-background reconstruction strategy. This effectively distinguishes and retains crucial foreground information for driving decisions (such as vehicles, pedestrians, and road signs) while filtering out redundant background tokens.

\subsubsection{Integrating Large Language Models}
End-to-end autonomous driving frameworks are capable of generating control commands directly from raw sensory inputs, yet they typically lack a deeper understanding of driving semantics and offer limited interpretability. Although large language models (LLMs) provide strong reasoning capabilities and natural language explanations that could enhance interpretability, their substantial computational overhead conflicts with the stringent real-time constraints of autonomous driving systems. Furthermore, a fundamental mismatch exists between the high-level textual reasoning performed by LLMs and the low-level trajectory planning required for vehicle control.
To overcome these challenges, DSDrive~\citep{liu2025dsdrive} introduces a lightweight end-to-end paradigm that distills the reasoning capabilities of a large vision-language model (VLM) into a compact LLM through Chain-of-Thought (CoT) knowledge distillation. Additionally, it proposes a waypoint-driven dual-head coordination module that unifies dataset structure, optimization objectives, and the learning process. Within this module, the reasoning head produces detailed reasoning steps and final decisions, while the planning head predicts trajectory waypoints essential for vehicle control. Both heads are jointly optimized in a unified framework, allowing the planning output to be directly grounded in the reasoning process and thereby enhancing interpretability and reliability. During deployment, DSDrive significantly reduces inference latency and memory consumption relative to full-scale VLM-based systems, providing a practical and computationally efficient solution for integrating language-driven reasoning into autonomous driving under constrained onboard resources.

\subsection{Experimental Comparison and Analysis}
% 汇总各类方法在公开数据集（如nuScenes, CARLA, Argoverse）上的性能对比
% 指标：mAP, FPS, 模型大小
% 表格形式展示，附简要分析
%分析要点：
%量化通常能提升推理速度，但可能牺牲精度,蒸馏方法往往在不增加推理开销的前提下显著提高轻量模型性能。结构剪枝和低秩分解的效果差，非结构化剪枝和稀疏化落地难，但在特定场景有效。对比不同方法的trade-off。

\begin{table}[!ht]
\centering
\caption{Experimental Comparison.
Results of PlanKD and InterFuser are reported from the original paper due to the complexity of end-to-end system reproduction.}
\renewcommand{\arraystretch}{1.15}
\scriptsize
\setlength{\tabcolsep}{3pt}
\begin{adjustbox}{max width=\textwidth}
\begin{tabular}{@{}cc>{\centering\arraybackslash}m{3.6cm}cccc@{}}
\toprule
\textbf{Task} & \textbf{Model} & \textbf{Backbone} & \textbf{Param.} & \textbf{FPS} & \textbf{mAP / Score} & \textbf{Dataset} \\
\midrule
\multirow{4}*{\begin{tabular}[c]{@{}c@{}}Object Detection\\(Quantization)\end{tabular}}
& PETR & ResNet50 & 35.8M & 12.23 & 0.341 & nuScenes \\
& Q-PETR & ResNet101 & 59.2M & 21.3 & 0.353 & nuScenes \\
& PETR & VoVNet & 81.66M & 19.33 & 0.40 & nuScenes \\
& Q-PETR & VoVNet & 81.66M & 34.75 & 0.41 & nuScenes \\
\midrule
\multirow{3}*{\begin{tabular}[c]{@{}c@{}}Object Detection\\(Distillation)\end{tabular}}
& DistillBEV & CenterPoint$\rightarrow$BEVFormer(R50) & 48.3M & 6.38 & 0.374 & nuScenes \\
& BEVFormer-Tiny & ResNet50 & 33.52M & 23.67 & 0.252 & nuScenes \\
& BEVFormer-Small & ResNet101 & 59.46M & 8.46 & 0.370 & nuScenes \\
\midrule
\textbf{Task} & \textbf{Model} & \textbf{Backbone} & \textbf{Param.} & \textbf{Inference
Time(ms)} & \textbf{Score} & \textbf{Dataset} \\
\midrule
\multirow{2}*{End-to-End}
& InterFuser & -- & 52.9M & 78.3 ms & 94.88 & CARLA \\
& PlanKD & InterFuser & 26.3M & 39.7 ms & 93.69 & CARLA \\
\bottomrule
\end{tabular}
\end{adjustbox}
\end{table}

This subsection presents an experimental comparison and analysis of representative model compression techniques applied to Transformer-based autonomous driving models. The evaluation focuses on two compression strategies that currently exhibit the highest practical applicability and reproducibility in the field: quantization and knowledge distillation. All experimental results reported in this paper are obtained using official configurations with publicly available and complete codebases as well as pre-trained models. For methods whose code repositories are incomplete or whose experimental settings do not allow reliable reproduction, we provide only qualitative analysis, without reporting quantitative results.

For Transformer-based 3D object detection quantization methods, this study selects PETR and Q-PETR as representative models to analyze the performance of quantization in multi-view 3D object detection tasks. As shown in Table IV, the baseline PETR model exhibits moderate inference efficiency. With a ResNet50 backbone, it achieves an inference speed of 12.23 frames per second (FPS) and a mean Average Precision (mAP) of 0.341. When using a VoVNet backbone, the inference speed increases to 19.33 FPS, accompanied by an improved mAP of 0.40. In contrast, the Q-PETR model, under an INT8 post-training quantization strategy, achieves a substantial improvement in inference speed without incurring noticeable accuracy degradation. Specifically, the Q-PETR model with a ResNet101 backbone reaches 21.3 FPS while maintaining an mAP of 0.353. The VoVNet-based variant of Q-PETR further improves inference efficiency to 34.75 FPS, with a slight increase in mAP to 0.41. These results demonstrate that quantization-aware design tailored to Transformer architectures is critical for maintaining numerical stability in 3D object detection models. For practical deployment in autonomous driving systems, task-specific quantization strategies can effectively leverage hardware acceleration while preserving detection performance.

We further examine the effectiveness of knowledge distillation in bird’s-eye-view (BEV) object detection models. As summarized in Table IV, DistillBEV is adopted as a representative distillation-based method, which transfers knowledge from a LiDAR-based teacher to a camera-based student model, which is instantiated as BEVFormer with a ResNet50 backbone. After distillation, the student model contains 48.3M parameters, achieves an inference speed of 6.38 FPS, and reaches an mAP of 0.374. For comparison, we evaluate standard BEVFormer baselines without distillation. BEVFormer-Tiny based on a ResNet50 backbone, achieves a high inference speed of 23.67 FPS, but suffers from a substantial drop in detection accuracy, attaining an mAP of only 0.252. In contrast, BEVFormer-Small, built on a ResNet101 backbone, improves detection performance to 0.370 mAP at the cost of reduced inference speed (8.46 FPS) and increased parameter count (59.46M). In general, the distilled model exhibits performance that lies between these two baselines. These results demonstrate that knowledge distillation can effectively mitigate the accuracy degradation caused by model lightweighting, enabling a more balanced trade-off between computational efficiency and detection performance in BEV-based object detection.

End-to-end autonomous driving models must satisfy the requirements of closed-loop control, and therefore impose much stricter constraints on inference latency than perception-only tasks. In this experiment, InterFuser is adopted as the teacher model, and the PlanKD distillation framework is evaluated on the CARLA simulation benchmark. Due to the high complexity of end-to-end driving systems and the substantial engineering cost required for faithful reproduction, the results reported here are directly taken from the original PlanKD publication. As shown in Table IV, the original InterFuser model contains 52.9M parameters, achieves a driving score of 94.88, and exhibits an inference latency of 78.3 ms. After applying knowledge distillation, the PlanKD model reduces the parameter count by approximately 50\%, to 26.3M. Meanwhile, inference latency is nearly halved to 39.7 ms, while the driving score decreases by only 1.19 points. These results demonstrate that task-oriented distillation strategies can substantially improve inference efficiency without compromising driving reliability.

\subsection{Emerging Compression Techniques and Future Directions}
%全整型量化技术，上个时代，逐渐淘汰。
%低秩分解（效果一般，落地难）
%误差补偿系列，GPTQ。主流
%Smoothquant+quarot
%FlashAttention系列。实验阶段但潜力巨大。
%未来展望，二值化，diffusion系列，MOE系列

As large-scale foundation models (e.g., Transformers and Mixture-of-Experts) become increasingly central to autonomous driving, the exponential growth in parameter scale has necessitated more aggressive compression strategies. Beyond traditional methods, recent research has explored advanced techniques such as error-compensated quantization, FlashAttention, and sparse MoE architectures. This subsection reviews these emerging trends and outlines future research directions.

Quantization is currently the most mature compression technique at the deployment stage. Early full-integer formats such as INT8 and FP8 have become standard on mainstream hardware, but traditional methods introduce significant nonlinear distortion when pushed to very low bit-widths ($\le$4 bits). This has led to rapid progress in error-compensation–based quantization. GPTQ~\citep{frantar2022gptq} uses a second-order Taylor expansion and a Hessian approximation to minimize reconstruction error, making it a dominant approach for post-training quantization. FOEM~\citep{zheng2025first} later introduced first-order gradient compensation and achieved notable improvements in low-bit accuracy. CDQuant~\citep{nair2024cdquant} combines greedy coordinate descent with optimal weight clipping (OWC) to provide better layer-wise convergence and stability. In addition, frameworks such as SmoothQuant~\citep{xiao2023smoothquant} and QLoRA~\citep{dettmers2023qlora} use channel scaling and mixed-precision quantization to maintain high accuracy while reducing latency.

Pruning, as a major direction for model acceleration, is generally divided into structured pruning and unstructured sparsification. Structured pruning removes entire layers, channels, or attention heads, directly reducing computational load. Methods such as ZipLM~\citep{kurtic2023ziplm} perform adaptive pruning through hardware-aware latency estimation and achieve real speedup on standard dense operators. However, the coarse granularity of structured pruning often leads to accuracy degradation if not followed by careful fine-tuning. Studies such as Sheared LLaMA~\citep{xia2023sheared} show that excessive pruning can cause “representation collapse,” suggesting that pruning should be combined with distillation or sparse reconstruction strategies. In contrast, unstructured sparsification removes individual weights and can reach very high compression ratios. Methods like SparseGPT~\citep{frantar2023sparsegpt} and Wanda~\citep{sun2023simple} achieve over 50\% sparsity without retraining, and when combined with quantization, as in Flash-LLM~\citep{xia2023flash}, can further improve throughput.

Low-rank decomposition is based on the assumption of linear redundancy and factorizes large matrices into products of low-rank matrices, reducing computation and storage while preserving continuous structure. Its main advantage is that it remains structured and compatible with dense operators, allowing the low-rank components to be executed efficiently. As a low-rank decomposition technique, SVD can in principle reduce memory and computation by truncating the singular values of a matrix and factorizing a large weight matrix into the product of two smaller ones. It also allows flexible control over the compression ratio and has shown strong effectiveness in areas such as image compression, suggesting potential for LLM compression~\citep{sharma2023truth}~\citep{hsu2022language}. However, existing SVD-based approaches for LLMs have shown limited success. Traditional methods directly truncate model weights, which leads to substantial performance degradation and requires extensive retraining.~\citep{sainath2013low}~\citep{lebedev2014speeding} As a result, low-rank decomposition is used more often for efficient fine-tuning, as in LoRA and DoRA. When applied directly for deployment, both compression and accuracy preservation are unsatisfactory; for example, ASVD~\citep{yuan2023asvd} shows that once the compression ratio drops below 0.9, accuracy loss exceeds 10\%, whereas quantization methods such as GPTQ remain nearly lossless at a 25\% compression ratio. Therefore, deployment with low-rank decomposition still heavily relies on fine-tuning, increasing operational costs.

Knowledge distillation transfers the implicit knowledge of a teacher model—such as logits, feature distributions, or task-alignment behavior—to a student model, allowing performance to be preserved during compression. Recent methods such as MiniLLM~\citep{gu2023minillm} have enabled effective cross-stage semantic transfer, allowing student models to maintain comparable accuracy while substantially reducing parameter counts. For Transformer-based autonomous driving systems, distillation is also extending toward multimodal semantic alignment. For example, using the BEV representations of a teacher model or the features of temporal attention to guide a lightweight student model can simplify the architecture while maintaining performance in tasks such as image–radar fusion or trajectory planning. Despite these benefits, distillation introduces additional training cost and often requires high-quality data. Its effectiveness also depends on the coverage of the teacher model and the design of the distillation loss. For instance, TAPIR~\citep{yue2024distilling} improves convergence through curriculum-based strategies, yet still faces limitations in domain generalization.

Even with compressed models, the quadratic complexity of the attention mechanism remains a fundamental runtime bottleneck. The FlashAttention family~\citep{dao2022flashattention,dao2023flashattention,shah2024flashattention} introduces block-level tiling to reduce memory traffic, effectively alleviating the O($N^2$) memory-access bottleneck and achieving 2--4$\times$ or greater speedups. FlashAttention-V2~\citep{dao2023flashattention} further reduces non–matrix-multiplication FLOPs, improving computational efficiency. The latest FlashAttention-V3~\citep{shah2024flashattention} applies hardware-aware optimizations tailored to modern GPUs and exploits asynchronous execution for scheduling and planning, pushing GPU utilization to a new level and delivering substantial performance gains. Theoretically, FlashAttention reframes attention from a compute-bound problem to an I/O-bound one: through tiling and streaming softmax, it transforms repeated HBM access for $QK^\top$, attention weights, and $V$ into a single block-wise read and write, keeping most data in on-chip SRAM before writing back the final results. Compared with methods such as pruning or quantization, which reduce HBM traffic through lossy compression, FlashAttention reduces bandwidth demands in a lossless manner.

While techniques such as FlashAttention focus on accelerating attention computation—addressing the question of how to compute faster—Mixture-of-Experts (MoE)~\citep{jordan1994hierarchical} architectures address a complementary challenge: how to increase model capacity without proportionally increasing inference cost. To fundamentally decouple model capacity from inference cost, Mixture-of-Experts architectures represent a structural paradigm shift. In MoE, a large network is divided into multiple expert sub-networks, and a gating network dynamically activates only a sparse subset of experts for each input. This enables the model to scale its knowledge capacity without proportional increases in FLOPs, thereby enhancing expressiveness while keeping inference efficient. Such a property is particularly valuable in autonomous driving, where handling long-tail scenarios demands extensive knowledge, but onboard hardware imposes strict real-time constraints. The CBDES MoE~\citep{xiang2025cbdes} framework exemplifies this potential by integrating structurally heterogeneous expert networks at the functional module level of the driving stack, combined with a lightweight Self-Attention Router gating mechanism for dynamic expert path selection. As the first modular MoE framework at this granularity in autonomous driving, CBDES MoE demonstrates the ability of sparse, input-aware inference to increase representational power without adding latency.

Model compression is therefore becoming a central enabler of practical autonomous driving deployment. Although software-level optimizations—such as lightweight network design and efficient attention mechanisms—can improve efficiency to some extent, current evidence suggests that they are often insufficient by themselves to satisfy real-time in-vehicle constraints without a measurable loss of capability. Consequently, compression techniques should be viewed not as optional engineering refinements, but as core components of deployable system design.
Among existing approaches, quantization, pruning, sparsification, low-rank decomposition, and knowledge distillation contribute from different angles and are complementary in practice. Quantization is especially attractive for resource-constrained deployment because of its strong hardware compatibility and favorable compression ratio. Structured pruning aligns better with common accelerators and can reduce both computation and memory bandwidth, whereas unstructured sparsification is appealing for extreme compression but often depends on specialized hardware or software stacks to translate sparsity into real speedups. Low-rank decomposition provides a continuous approximation route, yet its direct deployment benefits remain less mature than those of quantization and distillation. Distillation, in turn, is valuable when preserving task-critical behavior is more important than maximizing raw compression, particularly in multimodal or planning-related settings.
Looking ahead, the most important research direction is not a single universally superior compression algorithm, but a tighter coupling between task demands, model architecture, compression strategy, and hardware target. Autonomous driving systems will increasingly require evaluation pipelines that jointly consider perception quality, control reliability, runtime, memory footprint, and safety under degraded or long-tail conditions. In this sense, compression should be integrated throughout model design, training, validation, and deployment, rather than appended after model development.

\section{Conclusion}
\label{Sec: Conclusion}
This survey reviewed representative Transformer-based models for autonomous driving across perception, prediction, planning, and end-to-end systems, with an emphasis on how their architectural strengths relate to deployment constraints. Beyond cataloguing models, we examined model compression and acceleration as practical mechanisms for reconciling Transformer capacity with the latency, memory, and energy budgets of onboard systems. The central message of this survey is that efficient autonomous driving is not achieved by architecture innovation alone; it depends on coordinated choices spanning representation design, task formulation, compression strategy, and hardware execution.

Despite rapid progress, substantial barriers remain before Transformer-based autonomous driving systems can be considered truly deployment ready. Major obstacles include high computational and memory cost, sensitivity to long-tail real-world conditions, inconsistent reporting of deployment metrics, and the added complexity introduced by multimodal fusion and language-augmented driving frameworks. Future work should therefore move beyond reporting task accuracy alone and place greater emphasis on standardized deployment benchmarks, safety-aware evaluation under compression, and tighter co-design across model architecture, compression strategy, and hardware platform. More broadly, the field would benefit from evaluation protocols that distinguish open-loop accuracy from closed-loop driving reliability and that explicitly quantify the risks introduced by aggressive model compression. Progress in these directions will be essential for autonomous driving systems that balance accuracy, efficiency, robustness, and interpretability in reliable real-world operation.

\bibliographystyle{cas-model2-names}
\bibliography{reference}

\end{document}